\documentclass[10pt,letterpaper]{article}

\usepackage{cogsci}

\cogscifinalcopy 

\usepackage[
    backend=biber,
    style=apa,
    natbib=true,
    doi=false,
    isbn=false,
    url=false,
]{biblatex}
\usepackage{pslatex}
\usepackage{float} 

\usepackage{amsmath}
\usepackage{stmaryrd}
\usepackage{amssymb}
\usepackage{graphicx}
\graphicspath{ {./graphics/} }
\usepackage{subcaption}
\usepackage{graphbox}
\usepackage{enumitem}
\usepackage{color,soul}

\title{Visual Theory of Mind Enables the Invention of Proto-Writing}
 
\author{{\large Benjamin A. Spiegel\normalfont{\textsuperscript{$\star$}}, \textbf{Lucas Gelfond}, \normalfont{and} \bf George Konidaris} \\
  Department of Computer Science, Brown University \\ $^\star$correspondence to \texttt{bspiegel@cs.brown.edu}}

\begin{document}

\maketitle

\begin{abstract}

Symbolic writing systems are graphical \textit{semiotic codes} that are ubiquitous in modern society but are otherwise absent in the animal kingdom. Anthropological evidence suggests that the earliest forms of some writing systems originally consisted of \textit{iconic pictographs}, which signify their referent via visual resemblance. While previous studies have examined the emergence and, separately, the evolution of pictographic systems through a computational lens, most employ non-naturalistic methodologies that make it difficult to draw clear analogies to human and animal cognition. We develop a multi-agent reinforcement learning testbed for emergent communication called a \textit{Signification Game}, and formulate a model of inferential communication that enables agents to leverage \textit{visual theory of mind} to communicate actions using pictographs. Our model, which is situated within a broader formalism for animal communication, sheds light on the cognitive and cultural processes underlying the emergence of proto-writing.

\end{abstract}

\section{Introduction}

\begin{figure}[t!]
\centering
\includegraphics[width=\linewidth]{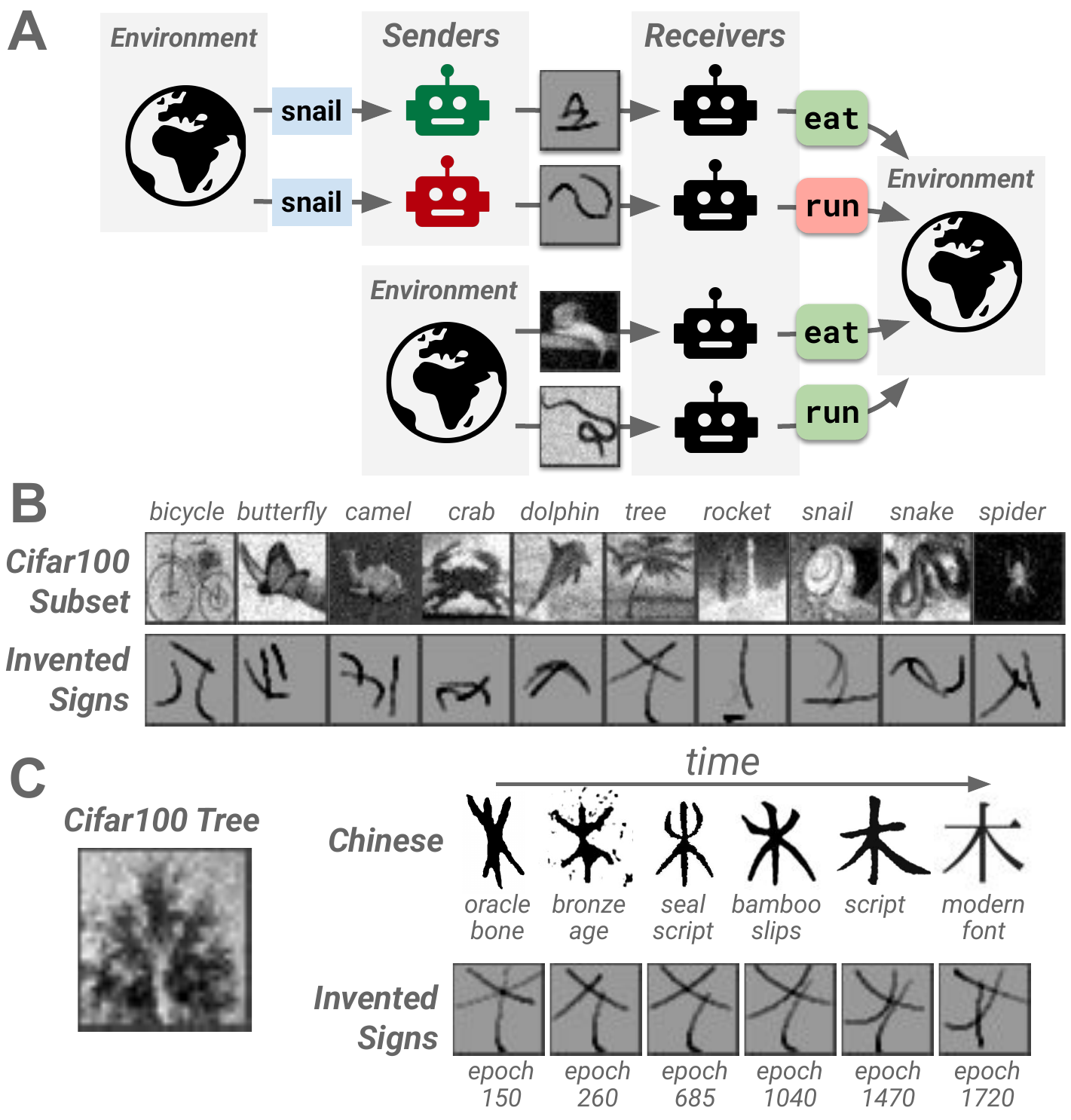}
\caption{\textbf{A)} 
Agents in a \textit{Signification Game} signaling for manipulative (red agent) or cooperative (green agent) ends. Actions yielding positive and negative rewards are denoted with green and red bubbles, respectively. Agents perceive signals and environment observations with the same networks.
\textbf{B)} Signs invented by agents to mimic classes from the Cifar100 dataset. Agents create pictographs given discrete states that serve as class labels. \textbf{C)} We simulate communication over long time horizons, finding that initially iconic signs grow more abstract over time. We compare the evolution of a sign for Cifar100 palm trees for a Chinese character for ``tree.''}
\label{fig:front-page}
\end{figure}

The ability to spontaneously invent, use, and understand natural language is unique to humans and separates us from the rest of the animal kingdom. However, the capacity for \textit{signification}, i.e. communicating a mental referent using a \textit{sign}, is shared with animals to some extent.
For example, sophisticated acoustic signaling has been observed in many species, including Sperm Whales \citep{sharma2024contextual}, Chickadees \citep{chickadee}, and the Japanese great tit bird \citep{birds}; and gesture-based visual signaling has been observed in chimpanzees, bonobos, and bees \citep{gillespie2014gesture, halina2013ontogenetic, von1962dialects}.

One variety of signification systems not found elsewhere in nature is the encoding of meaning in a visual medium using durable marks, e.g. \textit{human writing}. While diachronic linguistics has long studied the evolution of spoken language, the early origins of written language are primarily a matter of prehistoric record.
Substantial archaeological evidence indicates that the earliest forms of human writing consisted of \textit{iconic pictographs}, which signify their referents via visual resemblance \citep{taylor1883alphabet, trigger1998writing}.\footnote{Some of these systems were fully \textit{semasiographic}, conveying meaning directly without encoding speech; others included a \textit{glottographic} component encoding spoken language. We refer to such systems broadly as \textit{proto-writing}.} 
These stand in contrast to the \textit{abstract symbolic} signs used in modern writing systems, which instead rely on codified convention to convey meaning. The evolution of writing from \textit{icons} to \textit{symbols} raises fundamental questions about the cognitive mechanisms that enabled the invention of the first pictographs and the cultural processes driving their transition to abstract symbols.

While studies using human subjects have shed light on the short-term evolution of ad-hoc graphical systems \citep{garrod2007foundations, fay2014iconicity, fan2018commonobject, hawkins2023visual, fan2023drawing}, current computational simulations treat the invention of pictographs and the evolution of graphical systems as separate phenomena and rely on technical assumptions that lack faithfulness to a naturalistic setting for communication. For example, \citet{mihai2021draw} present a two-agent model for sketching that is end-to-end differentiable, enabling gradients to pass between agents that represent separate entities; \citet{fan2019visualabstraction} present a sketching model that is not generative, and instead communicates by selecting a sketch from a dataset of human-drawn sketches; \citet{qiu2022conventions} pre-train a sender model on an existing sketch dataset and use a hand-designed architecture that extracts salient edges for easier sketching; and \citet{jiang2024findingstructurelogographicwriting} model the evolution of Chinese using a domain-specific language with primitives that are crafted specifically for Chinese, and also do not consider how the first Chinese characters emerged.
These methodological limitations often make it difficult to draw clear analogies to human cognition, and fall short of situating \textit{pictographic signification} within a broader formalism for animal communication,
which could shed light on why even our closest relatives in the clade of Great Apes did not develop proto-writing and cannot understand iconic pictographs as well as humans can \citep{persson2008}.

We present a unified computational approach to simulating \textit{both} the invention of iconic pictographs and their transition to abstract symbols over time. In Part I we introduce the \textit{Signification Game}, a naturalistic communication testbed inspired by referential and signaling games \citep{lewis1969convention, skyrms2010signals} in which agents engaged in a decision process communicate by painting splines on a canvas, an action space designed to mimic that of proto-writing. In contrast to referential games, agents lack specialized hardware for processing communicative signals, and must instead learn to interpret signals using the same perception system used for processing state observations.
We show that while agents can learn to communicate from reward-maximization alone,
such a means of language learning is severely limited. Namely, the constraints imposed by a writing-like signal induces a \textit{signification gap}, making some referents extremely difficult to communicate without substantial learning by listeners.
In Part II, we find experimental evidence for such signification gaps, and formalize an inferential model of communication that can surmount signification gaps by leveraging \textit{visual theory of mind}, enabling communication of referents on short time scales. We further introduce the notion of \textit{referent sensitivity}, which partially compensates for the difficulty of communicating complex referents with a simple signaling space, leading to significant improvements in communication.

\section{Part I: A Model of Primitive Communication}

We describe a setting for simulating the emergence of graphical communication called a \textit{Signification Game}.
We start by drawing from theories on the emergence of primitive communication in nature to motivate our study design.

\subsubsection{The Origins of Communication in Nature} 

One plausible origin story for animal communication is that primitive communication acts constituted the use of signals to induce behavior in others that conferred an advantage to senders. For example, an animal that has associated a screeching noise with the presence of a predator may have learned to reliably take a fleeing action, a behavior that can be exploited by another animal who screeches so it may selfishly engorge itself on a scarce food source without competition. 
Formally, an animal who forms a stimulus-response (S-R) behavior is conditioned to associate a stimulus $S$ (e.g., a screeching sound) with a response $R$ (e.g., a fleeing action) using a classifier $f_\text{S-R}:X \to [0,1]$, where $X$ is the space of possible stimuli and $f_\text{S-R}(S)$ outputs the probability that $S$ is recognized as the stimulus and triggers the associated response behavior \citep{skinner1938behavior}. Such an S-R behavior can be exploited by another animal that can generate $\Hat{S}$, a convincing replica of stimulus $S$ that fools $f_\text{S-R}$ into inducing response $R$. Under this formulation, a primitive language can emerge comprised of utterances $\{\Hat{S}_1, \Hat{S}_2, ...\}$ with meanings co-opted from precursor S-R behaviors $\{R_1, R_2, ...\}$.\footnote{A similar action-oriented account of language is described in \citet{philosophicalinvestigations}.}
The communication-as-manipulation hypothesis is attractive because it suggests that communication can emerge without any special-purpose cognition; as soon as biological hardware that can generate signals emerges, the same cognitive tools for maximizing reward in the environment can be applied to such hardware to learn primitive languages.

\subsection{Methods}

\subsubsection{Signification Games}

A Signification Game is an iterated multi-agent reinforcement learning \citep{littman1994markov} task for simulating the emergence of communication under naturalistic assumptions. A population of agents begins by learning S-R behaviors---implemented as policy functions from observations to actions, $\pi:O \to A$---that maximize reward in an underlying decision process. Over time, agents are gradually given the opportunity to interact, allowing them to communicate using a continuous signal---in our case, greyscale images---that is interpreted by the same behavioral policy used for interacting with the environment. In its most basic formulation, agents cannot distinguish between their observations of the world and the signals generated by other agents; agents must learn to exploit the behavioral policies of others to induce desired actions, which may be selfish or mutually-advantageous.
Signification Games can be played over millions of rounds to give us a picture of how communication protocols emerge and evolve over time.

Formally, a Signification Game is a decentralized partially-observable Markov Decision Process \citep{oliehoek2016decpomdp} denoted as a tuple \( \langle S, \{ A_i \}_{i=1}^{N}, \{ \Omega_i \}_{i=1}^{N}, T, R\rangle \), where
\( S \) is a set of states,
\( A_i \) is the action space for agent \( i \),
\( \Omega_i (o_i| s)\) is the observation function for agent \( i \),
\( T(s'|s, \{ a_i \}) \) represents the transition function given the current state \( s \) and the joint actions \(\{ a_i \}\),
and \( R(s, \{ a_i \}) \) is the reward function. From each state $s \in S$, a collection of contextual bandits\footnote{
Using a contextual bandit simplifies credit assignment, though our methods extend to temporally-extended decision processes.} $\{c_1, ..., c_n\}$ \citep{langford2007bandit} with winning arms $\{a_1, ..., a_n\}$ is created for a subset of agents designated as \textit{receiver} agents. In our simulations, the contexts $\{x_1, ..., x_n\}$ for each bandit are greyscale images drawn from two sources: 1) a multi-class dataset whose classes correspond to the winning arms, or 2) from \textit{sender} agents who observe the winning arms before generating a signal. This gives agents the opportunity to induce actions in others based on knowledge of the underlying decision process.

\subsubsection{Agents}

Agents in a Signification Game have a high-level policy $\pi$ comprised of two sub-policies: a behavioral policy $\pi_R$ invoked when agents play a \textit{receiver} or \textit{observer} role and a sender policy $\pi_S$ invoked when agents play a \textit{sender} role. The behavioral policy acts directly on the environment, taking images as input---either observed from the environment or generated by other agents---and outputting discrete actions to the underlying decision process. The sender policy generates an image conditioned on the discrete state of the underlying decision process. Formally, each agent $i$ seeks to maximize cumulative returns using its high-level policy $\pi_i$ over the duration of the Signification Game: $\mathbb{E} \sum_{t=0}^\infty \gamma^t R(s_t, \pi_i(\Omega_i(s_t))),$
where sub-policies are conditionally deployed based on the observation function $\Omega_i:s_t \mapsto o_i$, which returns an image if the agent is tasked with \textit{observing} or \textit{receiving}, or a one-hot state vector if the agent is tasked with \textit{sending}.

Behavioral policies are implemented as Convolutional Neural Networks \citep{alexnet} with an attached neural network, whose logits are akin to the activations for a set of S-R classifiers $\{f_{a}|a \in A\}$ from which actions are sampled according to a categorical distribution obtained by applying a softmax: $P(a|o) = \text{softmax}(\{f_{a_i}(o)\})$. Sender policies are implemented with neural networks that output a multivariate Gaussian distribution over spline parameters, which are sampled and rendered as an image. Sender policies are akin to a set of generators $\{g_{a}|a \in A\}$ for producing stimuli that induce actions in listeners. Policies are trained independently with PPO \citep{schulman2017ppo} using separate rollouts $\{(s_t, \pi_S(o_i^t), r_t), ...\}$ and $\{(s_t, \pi_R(o_i^t), r_t), ...\}$. Our agents use a signaling space akin to that of human proto-writing, however, Signification Games can potentially be used as a general model for communication by adjusting observation and action spaces to reflect various communication mediums.

\begin{figure*}[t]
\centering
\includegraphics[width=0.96\linewidth]{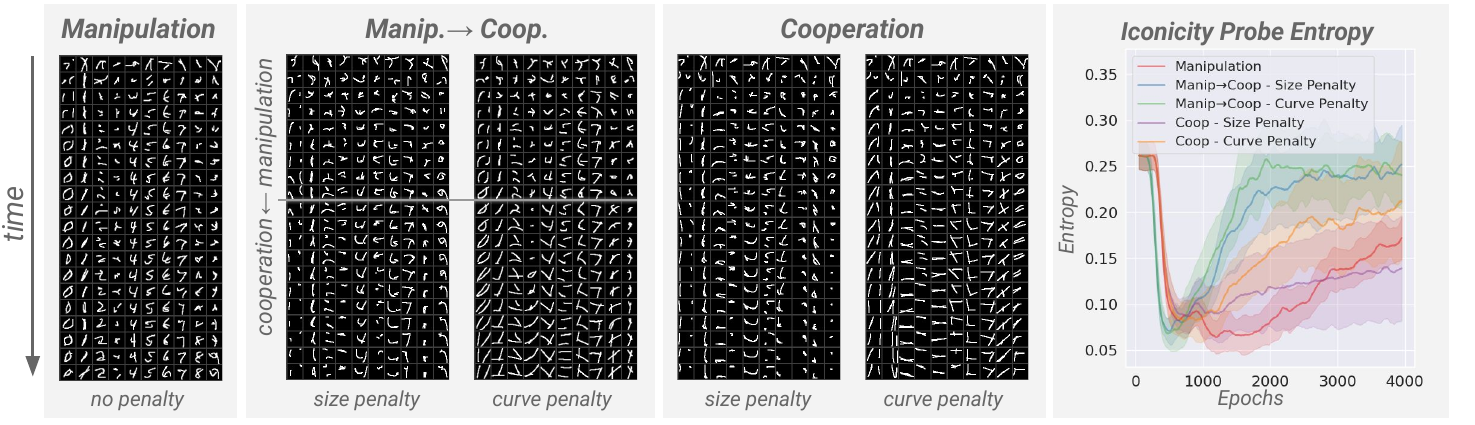}
\caption{\textbf{Visualization of Signaling Over Time}. Agent signals first appear random (top rows) and grow increasingly ordered over time. In a manipulation setting (left), senders learn to generate convincing replicas of state observations, tricking increasingly picky listeners. In a manipulation-cooperation setting (second from left), agents begin cooperating after 300 epochs of manipulation, resulting in a drift from iconic signals into abstract symbols. In a fully-cooperative setting (second from right) agents generate vaguely iconic signals that quickly shift into symbols. Environmental pressures on signal production and interpretation---simulated by size and curve penalties---greatly affect the appearance of signals over time, as evidenced by the final rows of signs. Qualitative analysis of the \textit{iconic} nature of signals is supported by an \textit{iconicity probe} (right, average of 10 runs plotted with a 95\% c.i.). We find that signals achieve low probe entropy (i.e. higher confidence of recognition) shortly after communication begins before increasing dramatically, reflecting a transition from icons to abstract symbols.
}
\label{fig:banner}
\end{figure*}

\subsection{Studies}

Signification Games afford a number of tunable parameters that can drastically affect simulation outcomes including the payoff structure for communication (e.g., cooperative or manipulative), the population size, the signal-rendering system for generating images, the frequency of agent-agent interaction over time, and the multi-class image dataset used in the underlying decision process. Unless otherwise stated, simulations use the following parameters.

\begin{itemize}[noitemsep,topsep=0pt]
    \item Each simulation begins with a solipsistic phase in which agents interact solely with the underlying decision process until performance plateaus, at which point agents are gradually exposed to signals generated by other agents. We imagine a setting in which the hardware for communication emerges in a population of agents who are already proficient at maximizing reward in their environment.
    \item Observations for the underlying decision process are drawn from the MNIST dataset \citep{mnist}. When observing images from the environment, agents are rewarded for taking a discrete action corresponding to the true observed image class. While agents essentially perform classification, this simulation is roughly analogous to animals learning to take advantageous actions in their environment, e.g. fleeing or pursuing, based on observed stimuli, e.g. from predators, competitors, or allies. We use the MNIST dataset due to its easily-recognizable classes, though a more natural image dataset is used in Part II.
    \item Senders output parameters for two Bézier curves that are rendered on a canvas and peppered with Gaussian noise to simulate an imperfect communication medium. Senders also receive rewards inversely proportional to the entropy of listener policies if successful communication occurs, reflecting agents with higher confidence in their observations being more likely to act on them. Effectively, this encourages senders to push away from the decision boundaries of receivers to generate more prototypical stimuli.
    \item Each simulation is run with a population of 5 agents. Simulations with up to 20 agents yield similar results, but using small populations enables longer simulations. In general, increasing the number of agents acts as a \textit{regularizer}, as agents must convince a larger population with their signals.
\end{itemize}
We additionally train an \textit{iconicity probe}: a CNN-based classifier trained on MNIST images to assess the iconicity of agent signals. The probe returns a distribution over MNIST classes for a given image. High entropy of this distribution indicates that the probe struggles to recognize the class of the image. We find that \textit{iconic} images yield lower entropy scores, reflecting higher confidence that the image belongs to a specific class, while more \textit{abstract symbolic} images yield higher entropy scores, reflecting more class ambiguity. \\

\subsubsection{Deceptive Signaling Leads to Iconic Replicas}

We first assess the emergence of communication in an adversarial context. In this study, senders are rewarded for successfully tricking receivers into taking an action, while receivers are penalized for reacting. As agents balance their capacity to classify environment observations while interpreting deceptive signals, an arms race unfolds in which agents generate increasingly convincing stimuli.\footnote{We note the similarity to a GAN \citep{goodfellow2014gan}.} As a result, agents learn to generate a form of iconic signs called \textit{replicas} (see Figure \ref{fig:banner}, left), which confer meaning by virtue of being perfect facsimiles of existing meaningful stimuli \citep{eco-semiotics}. Crucially, iconic signification using replicas is possible only when the sender signaling space directly overlaps with relevant environment stimuli. In Part II we eliminate this possibility.

\subsubsection{Cooperative Communication Leads to Abstract Symbols}

In contrast to adversarial signaling, receivers in a cooperative setting have an incentive to communicate and should share the effort of signal transmission by assuming an interpretive burden. While initial signals are likely to be iconic---so that their referents may be recognized by receivers---the eventual result of cooperative communication should be cost-minimizing signals.

We run simulations under two conditions: one in which agents begin cooperating after a period of adversarial communication providing them with a set of strongly-iconic signals, and another in which their first use of signals is cooperative. In both conditions we simulate a scenario where communicating with others is more lucrative than interacting with the environment directly, reflecting a population of agents with specialized roles that relies on communication to survive. As the proportion of agents who interact with the environment decreases, agents become the primary sources of observed stimuli,
alleviating pressures on senders to create signals resembling environment observations. The result is the collapse of initially iconic signals into \textit{abstract symbols} that drift while continuing to serve as reliable signifiers for referent behaviors.
Furthermore, as agents grow proficient at communicating, the initial pressures to merely cross receiver decision-boundaries cease to dominate the reward landscape, and environmental constraints that influence the perception and creation of signals become exaggerated, e.g. as constraints imposed by writing implements and mediums are hypothesized to have impacted the look of many insular Asian scripts \citep{miller2014survey}. We simulate these pressures using two signal penalties applied to senders: a \textit{curvature} penalty, which rewards agents for drawing straighter lines, and a \textit{size} penalty, which penalizes agents for larger signals.

\section{Part II: A Model of Pictorial Signification}

\begin{figure}[t!]
\centering
\includegraphics[width=0.80\linewidth]{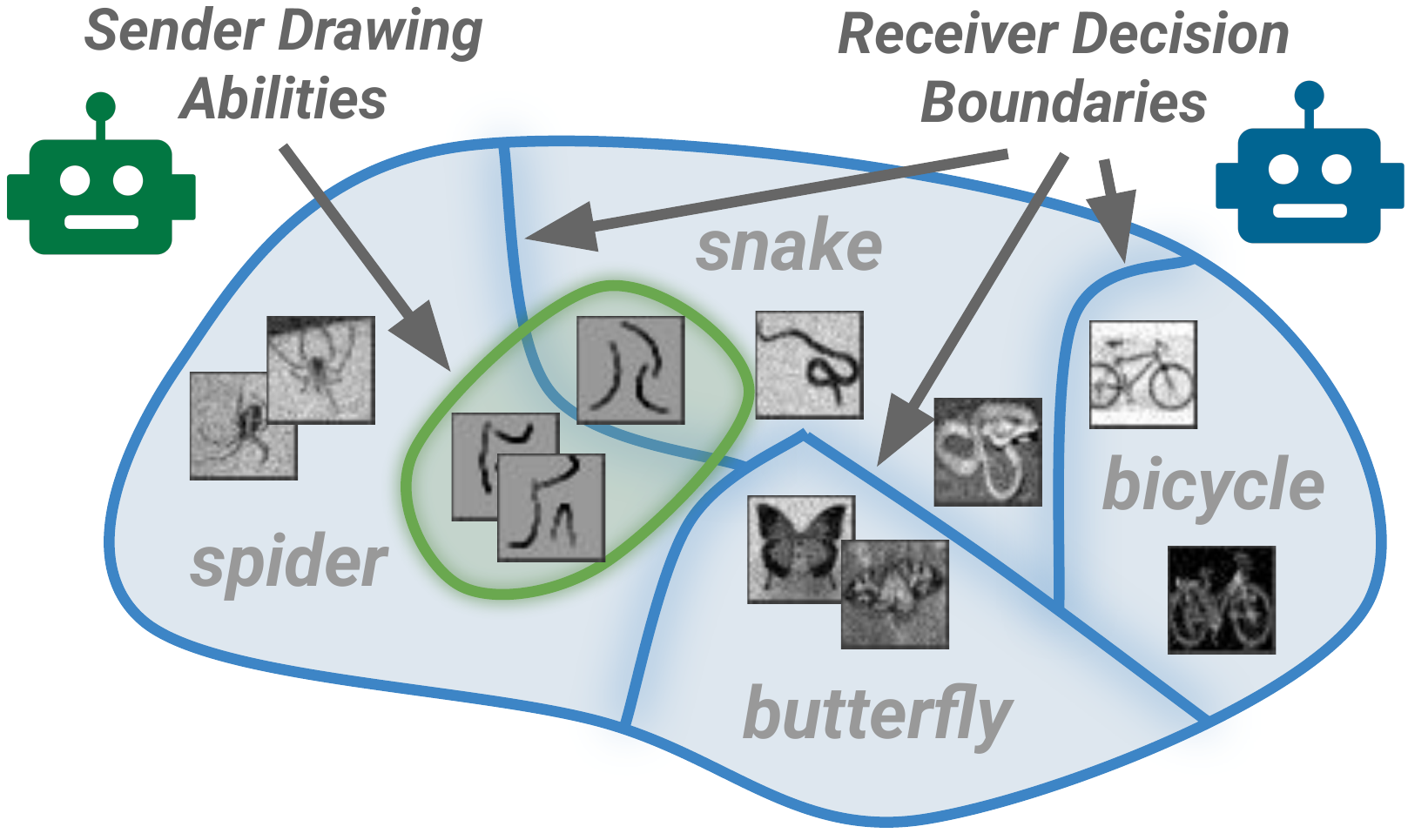}
\caption{Senders have difficulty generating convincing environment observations for most referents. The distance between the space of drawings and the decision boundary for a referent represents a \textit{signification gap} for that referent.}
\label{fig:sig-gap-diagram}
\end{figure}

In Part I, we demonstrated that primitive communication can spontaneously emerge once agents become capable of emitting stimuli, even when the method of processing observed stimuli remains the same. However, languages learned by such means are significantly limited by what they can reference. For a sign-referent pair to join the lexicon, senders must induce referent behaviors by subjecting receivers to convincing stimuli. If the stimuli for existing S-R behaviors is complex relative to the signaling medium---as was the case for early human proto-writing---agents may be unable to generate convincing replicas, leaving them unable to communicate. We refer to this barrier to communication as a \textit{signification gap} (see Figure \ref{fig:sig-gap-diagram}), and suggest that new methods for generating and processing stimuli are necessary to bridge this gap.

\subsubsection{Models of Inferential Communication}

Over the past decade, significant advancements have been made in formal accounts of \textit{inferential} communication, wherein receivers glean the meaning of signals by inferring the intentions of senders based on context and shared knowledge \citep{degen2023rsa}. Most notably, \citet{goodmanfrank2016rsa} introduced the Rational Speech Acts (RSA) framework, which formalized Gricean maxims \citep{grice1975-GRILAC-6, sperber1986relevance} by modeling communication as a recursive process where agents infer intentions based on models of their interlocutor. Similar Bayesian inference models have previously been introduced to infer the goals of agents based on their behavior \citep{ullman2009inverse, baker2009inverse} and more recent ``inverse-inverse" planning methods \citep{chandra2023acting} have leveraged such models to guide the production of new behavior in a manner akin to \textit{acting}.

At the core of these inferential models of communication lie conventional relationships between the actions of an agent and its intended goal,\footnote{This is formalized in RSA by a utility function, $U(u;w)$, based on a denotational semantics
providing the truth-value of a sentence in a given world. In inverse- and inverse-inverse planning it is formalized by a set of optimal Q-functions, $\{Q^{g_i}(s,a)\}$, capturing the degree to which behaviors are in service of a set of goals.}
leaving open questions about how such relationships \textit{become} conventionalized.
We imagine a setting in which a population of agents has not yet established graphical codes, i.e., there are no pre-existing conventional measures for relating drawn signs to their referents. We therefore hypothesize that to leverage inferential communication in order to signify visually-complex referents using simple marks, agents must \textit{invent} a graphical convention based on a mutual assumption about their lowest common denominator: convergence to similar models of perception and action that arise from interaction with a shared environment.

\subsubsection{Pictographic Signification with Visual Theory of Mind}

We seek to formalize the uniquely human capacity for \textit{pictographic signification}. Recall that pictographs are effective not because viewers are genuinely convinced they are observing the referent,\footnote{A pictographic horse is unlikely to convince you it is a real horse.} but because they consider the space of possible referents when interpreting them.
Interpreters of pictographs ask, ``What does this pictograph best resemble, given what it could refer to and how it otherwise could have looked?'' while creators of pictographs ask, ``How can I best show likeness to my target referent while showing distinctness from other referents, given what I can draw?'' By leveraging such \textit{visual theory of mind} to reason about the drawing abilities and perception systems of others, humans can communicate referents using even crudely-drawn iconic pictographs. We formalize this model of communication as:
\begin{align*}
    P_{send}(s|r_i) \propto \frac{f_{r_i}(s)}{\sum_{j \neq i} f_{r_j}(s)}&,\ \
    P_{recv}(r|s) \propto P_{send}(s|r)P(r)P_{ref}(r), \\
    P_{ref}(r_i) &= \frac{1}{\int_S f_{r_i}(s)p(s) \ ds}.
\end{align*}
Senders select pictographs that maximally trigger the S-R classifier for a target referent and minimally trigger other classifiers,\footnote{This notion is similar to \textit{contrastive loss} \citep{hadsell2006contrastive}.} while receivers apply Bayes' rule to infer the likely referents of pictographs. Receivers are additionally tempered by a \textit{referent sensitivity} term, $P_{ref}(r_i)$, for capturing the overall likelihood of recognizing any pictograph as a given referent. This term is larger for referents with stimuli that are harder to replicate using pictographs, sensitizing receivers to referents with a greater signification gap.

\begin{figure*}[t]
\centering
\includegraphics[width=1.0\linewidth]{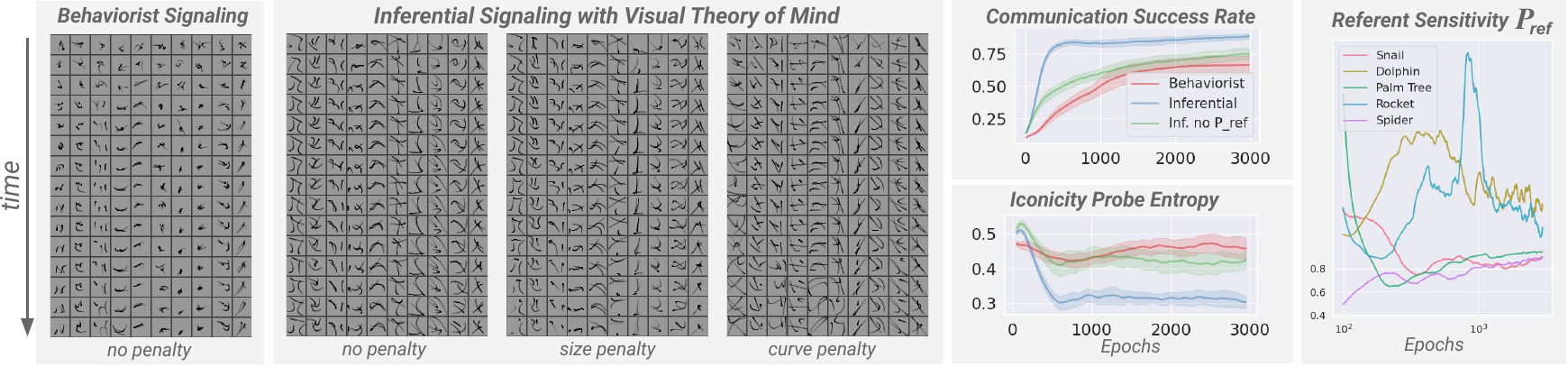}
\caption{\textbf{Comparison of Behaviorist and Inferential Signaling}. Agents wielding pictographic signification quickly succeed at communicating by generating iconic stimuli, achieving high rates of communicative success after only 300 epochs (second from right). As in Part I, we find that environmental pressures affect the appearance of pictographs over time (second group from left). We also report one agent's estimate of $P_{ref}(r_i)$ for five referents over time (right). Some $P_{ref}$ values for a small number of referents are volatile (e.g. dolphins and rockets) before drifting back toward the mean. We hypothesize this is due to the mastering of some referents before others, likely from inherent difficulties in creating iconic signs for some referents. Analysis of Cifar100 finds that there is greater visual variance in the classes for which $P_{ref}$ is more volatile (e.g. rockets are sometimes obscured by plumes of exhaust smoke).}
\label{fig:banner-part2}
\end{figure*}

\subsection{Methods}

\subsubsection{Implementing Theory of Mind with Mirror Cognition} The inferential communication model described above relies on two critical assumptions: 1) that agents have shared access to the same S-R classifiers, $\{f_{r_i}, ...\}$, when computing $P_{send}(s|r_i)$ and $P_{ref}(r_i)$; and 2) that agents can explore the entire space of pictographs to compute the integral in $P_{ref}(r_i)$. Neither are reasonable in a naturalistic setting where communication occurs between agents with siloed cognitive systems. Instead, we implement agents that leverage models of their own cognition as if they were models of their interlocutor \citep{gallese1998}, prompting agents to generate and interpret their own pictographs to estimate $P_{send}(s|r_i)$ and $P_{ref}(r_i)$. Before sending, agents generate a small number of pictographs and probabilistically select the one that maximizes $P_{send}$.

\subsection{Studies}

In a final set of studies we examine the inferential capacities of agents wielding pictographic signification. Simulations in this section use the following parameters:

\begin{itemize}[noitemsep,topsep=0pt]
    \item Observations from the underlying decision process are drawn from 5000 samples across 10 classes from the Cifar100 dataset \citep{krizhevsky2009learning} and converted to greyscale (see Figure \ref{fig:front-page}). A more naturalistic dataset is chosen to demonstrate how communication emerges when there are large signification gaps. A separate iconicity probe is trained on this dataset.
    \item Senders now generate parameters for 3 splines, and the sender canvas background color is adjusted to be closer to that of the average Cifar100 image.
    \item Each simulation is run with a population of 10 agents, which can now differentiate between environment observations and agent-generated pictographs, and use their inferential communication model only when interacting with other agents. Half of all receiver observations are generated by agents, and half from the environment.
\end{itemize}

\subsubsection{Proto-Writing Implements Induce Signification Gaps} An analysis of the referent sensitivities, $P_{ref}(r_i)$, for one agent finds that the constraints of its signaling space make some referents significantly harder to signify than others (see Figure \ref{fig:banner-part2}, right). Specifically, it finds it difficult to recognize the early scribblings of senders as being palm trees, as evidenced by a large $P_{ref}$, and is most likely to recognize them as spiders, as evidenced by a small $P_{ref}$. Behaviorist agents struggle to communicate in this setting, sending abstract symbols that do not resemble environment observations, as evidenced by high entropy in its iconicity probe evaluations. Behaviorist agents learn to communicate very slowly, potentially due to substantial efforts of receivers to learn \textit{indexical} relationships between communicated symbols, actions, and rewards.

\subsubsection{Inferential Communication Bridges Signification Gaps} While behaviorist agents struggle to communicate for 1000+ of epochs, inferential agents can successfully communicate after only 300 epochs (see Figure \ref{fig:banner-part2}, second from right, top). These agents are able to compensate for their \textit{prima facie} impressions of signals with the help of the $P_{ref}$ term, which down-weights referents with easily-recognizable stimuli and up-weights referents whose stimuli are more difficult to replicate. As agents grow more proficient at signaling, $P_{ref}$ values converge to the same value, signifying increasing ease in producing and interpreting signs for all referents. An ablation of the $P_{ref}$ term shows that referent sensitivity significantly impacts communication abilities, with the ablated model performing only moderately better than the behaviorist model.

\section{Discussion}

We set out to shed light on the mystery of why graphical proto-writing systems are unique to humans despite the ubiquity of symbolic communication in nature \citep{rosenthal2007} and the ability for some animals to learn abstract sign systems in structured learning settings \citep{rumbaugh1974lana, PATTERSON197872}. These phenomena suggest that the capacity for symbolic communication alone is likely not sufficient for the invention of proto-writing, which differs in form from the signaling systems already used by many communicative animals. Furthermore, the presence of visually-recognizable features in the earliest forms of human proto-writing suggest that abstract visual resemblance played a critical bootstrapping role in the adoption and invention of writing systems. We therefore suggest that this capacity---to leverage competence about one's own visual system and a space of possible referents when interpreting the intended referent of an iconic pictograph---could be a unique feature of human cognition that enabled early humans to invent the first writing systems.

We assess this hypothesis by analyzing emergent languages invented by groups of reward-maximizing agents engaged in \textit{Signification Games}, extensions of signaling games that situate agents in an underlying decision process with partial observability that can be bridged with communication. We first implement a behaviorist model of communication akin to animal signaling and identify a critical shortcoming: agents struggle to communicate most referents due to a \textit{signification gap} between their own capacity to signal and the stimuli needed to evoke referent behaviors. To surmount signification gaps, we formalize a model of \textit{pictographic signification} inspired by Bayesian inference that leverages \textit{visual theory of mind}, enabling the communication of most referents with crude pictographs that resemble early forms of human proto-writing.
We additionally analyze the evolution of such pictographs and find that they grow visually-abstract over time, taking on a similar trajectory to some ideographic writing systems invented by humans. Altogether, our results suggest that behaviorist models of communication are insufficient at explaining the emergence of writing systems, and that visual theory of mind may have played a crucial role in the invention of the first pictographic sign systems.
We believe these results warrant further investigation into the extent to which other animals have the theory of mind necessary for performing signification in various media, and we note the potential for leveraging Signification Games as a general model for communication over such media.

\section*{Acknowledgments}

This research was partially funded by the ONR under the REPRISM MURI N000142412603 and Grant \#00014-22-1-2592, as well as the National Science Foundation Graduate Research Fellowship Program under Grants \#2439559 and \#2040433. We thank Kartik Chandra, Aaron Kirtland, Daniela Mihai, Shane Parr, and James Tompkin for their thoughtful comments on this research.

\printbibliography

\end{document}